\documentclass[]{interact}

\usepackage{epstopdf}
\usepackage[caption=false]{subfig}
\usepackage[longnamesfirst,sort]{natbib}
\bibpunct[, ]{(}{)}{;}{a}{,}{,}

\usepackage[utf8]{inputenc}
\usepackage{hyperref}
\usepackage{booktabs}
\usepackage{multirow}
\usepackage{float}

\theoremstyle{plain}

\theoremstyle{definition}

\theoremstyle{remark}

\begin{document}

\articletype{Article}
\title{Using Particle Swarm Optimization as Pathfinding Strategy in a Space with Obstacles}
\author{
\name{David\textsuperscript{a} and Budi Adiperdana\textsuperscript{a}\thanks{CONTACT David. Email: david18001@mail.unpad.ac.id}}
\affil{\textsuperscript{a}Department of Physics, Faculty Mathematics and Natural Sciences, Universitas Padjadjaran}
}
\maketitle

\begin{abstract}
Particle swarm optimization (PSO) is a search algorithm based on stochastic and population-based adaptive optimization. In this paper, a pathfinding strategy is proposed to improve the efficiency of path planning for a broad range of applications. This study aims to investigate the effect of PSO parameters (numbers of particle, weight constant, particle constant, and global constant) on algorithm performance to give solution paths. Increasing the PSO parameters makes the swarm move faster to the target point but takes a long time to converge because of too many random movements, and vice versa. From a variety of simulations with different parameters, the PSO algorithm is proven to be able to provide a solution path in a space with obstacles.
\end{abstract}


\section{Introduction}

Swarm intelligence (SI) is an artificial intelligence approach that aims to solve problems using algorithms based on the collective behavior of social animals. Swarm intelligence is different from evolutionary algorithms. This is because the evolutionary algorithm develops a new population for each generation/iteration, while the swarm intelligence algorithm improves the individual for each generation/iteration \citep{bansal2019particle}. The most popular SI methods include Artificial Bee Colony Algorithm, Ant Colony Optimization, and Particle Swarm Optimization \citep{karkalos2019swarm}.

Particle Swarm Optimization (PSO) algorithm, which was invented by Kennedy and Eberhart \citep{kennedy1995particle}, is an optimization method inspired by the social behavior of animal such as bird flocking and fish schooling. PSO is a population-based search algorithm, where each individual/particle changes its position each time to explore a multi-dimensional space. The PSO algorithm combines the local search with the global search method. Each particle has a position and velocity in the space of a certain dimension. Initialization of the PSO algorithm starts by setting the initial position of the particle, and then looking for the optimal value by updating its position randomly. For each iteration, particle updates its position following the two best values: the best solution that has been obtained by itself (local best) and the best solution in the population (global best). After the particle position is evaluated by the fitness function, each particle "communicate" the information of best position to the other particles, so they can adjusts the position and velocity toward that best point.

\begin{figure}[ht]
	\centering
	\includegraphics[width=8.5cm]{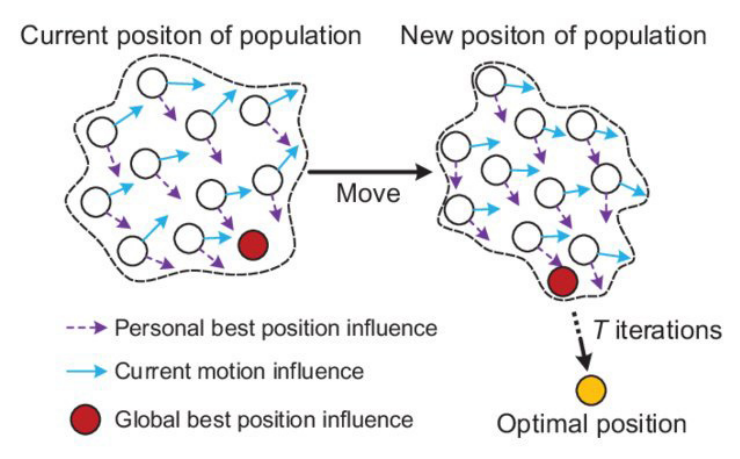}
	\vspace{-0.4\baselineskip}
	\setlength{\belowcaptionskip}{-5pt}
	\caption{\centering Particle position update process \citep{chen2020optimization}}
	\label{fig:update}
\end{figure}

One of the most important tasks in mobile robot navigation problems  is to plan a collision-free route from the starting point to the target position. One of them uses an optimization process. However, traditional optimization techniques create local minima, and the robot may not reach its goal even if a solution exists \citep{falco2020path}. Over the last few years, several heuristic algorithms have been introduced into path planning \citep{zhu2021new}, including PSO. The PSO algorithm is known to be very good for use in free space problems \citep{psosafril}. This paper aims to investigate the ability of PSO algorithm to find solutions/pathfinding in a space with obstacles, so if it is successful, it can be used to find solutions for maze games and path planning in robot movement.

\section{Algorithm}
With the standard PSO algorithm, the path planning can be divided into the following steps, as shown in Figure 2.

\begin{figure}[ht]
	\centering
	\includegraphics[width=14cm]{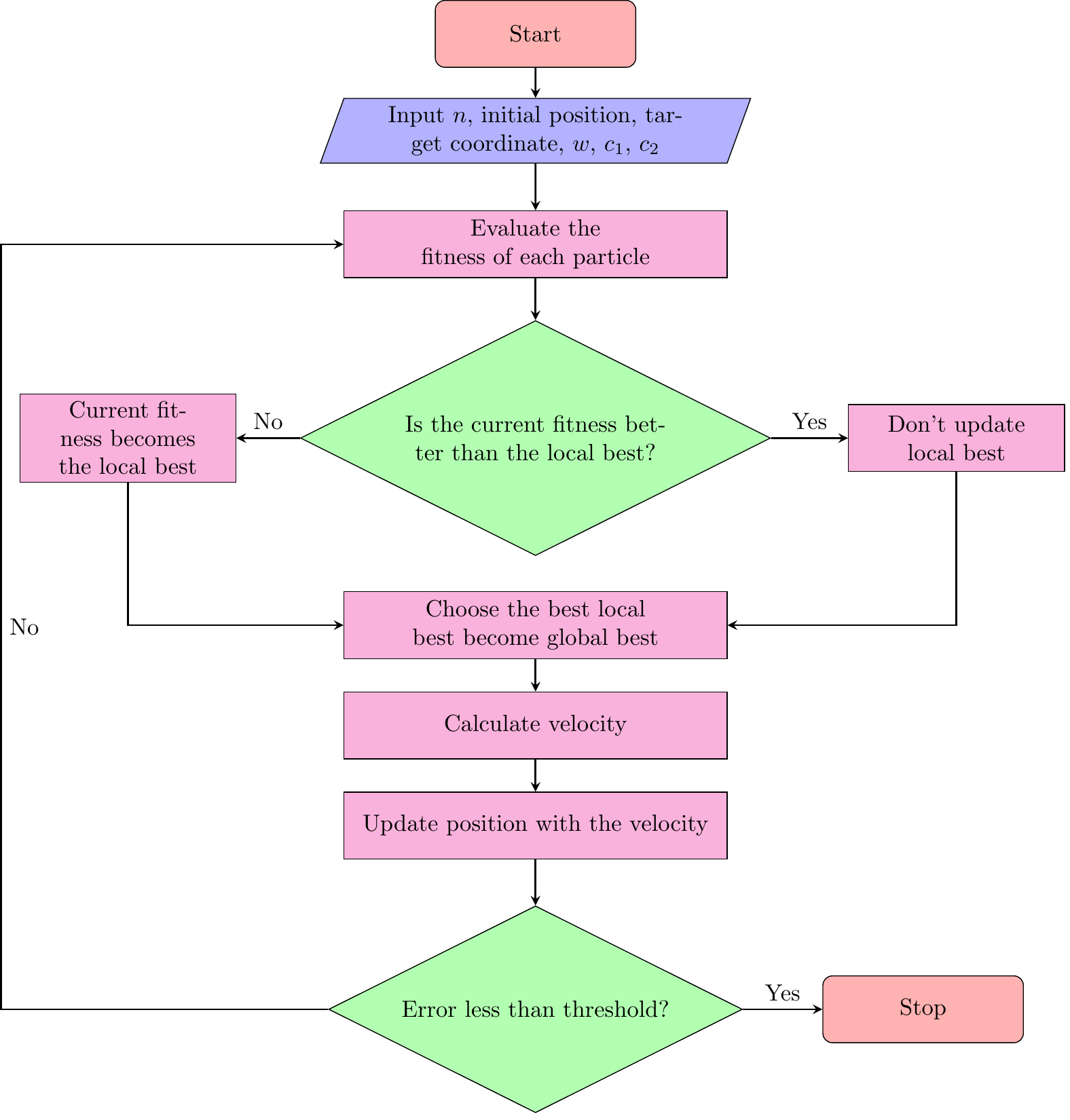}
	\vspace{-0.4\baselineskip}
	\setlength{\belowcaptionskip}{-5pt}
	\caption{\centering PSO Algorithm Flowchart}
	\label{fig:swarm}
\end{figure}

First step is initialize a starting position, target position, obstacle coordinate in the search space. Then, for each particle $i$, update the position of particle $i$ according to equation (1). 

\begin{equation}
	v_{i+1}=w \cdot v_{i}+c_{1} \cdot \operatorname{rand}_{1}\left(x_{best, i}^{P}-x_{i}\right)+c_{2} \cdot \operatorname{rand}_{2}\left(x_{best, i}^{G}-x_{i}\right)
\end{equation}\\
\noindent
where $v_i$ is the speed of the $i$-th iteration, $x_i$ is the particle position of the $i$-th iteration, $x_{best, i}^{P}$ is the local best position of the $i$-th iteration, $x_{best, i}^{G}$ is the globally best position of the $i$-th iteration, $w$ is the inertia weight of the current speed when updating the speed, $c_1$, $c_2$ are the follow factors, and rand$_1$, rand$_2$ are uniform random numbers from 0–1. After that, update the velocity of particle $i$ according to equation (2).

\begin{equation}
	x_{i+1}=x_{i}+v_{i+1}
\end{equation}

\noindent
Then evaluate its fitness value according to equation (3).

\begin{equation}
	f=\left(x_{\text {target}}-x\right)^{2}
\end{equation}\\
\noindent
The fitness function is used to evaluates how close current position is to the targeted position. Next, update $x_{best, i}^{P}$ and $x_{best, i}^{G}$ if necessary according to equation (4 \& 5). 

\begin{equation}
	\text x_{best, i+1}^{P} = \begin{cases}\text x_{best, i}^{P} & \text { if } f\text (x_{best, i}^{P}) \leq f\left(x_{i}\right) \\ x_{i} & \text { if } f\text(x_{best, i}^{P})>f\left(x_{i}\right)\end{cases}
\end{equation}
\begin{equation}
	\begin{aligned}
		& x_{best, i+1}^{G} =\min \{f(y), \quad f(x_{best, i}^{G})\} \\
		&\text {where,} \quad y \in\left\{x_{best, 0}^{P}, x_{best, 1}^{P}, ..., x_{best, N}^{P}\right\}
	\end{aligned}
\end{equation}\\\\
\noindent
If it has converged, the path is saved and the iteration is exited. Otherwise, it started again to continue the iteration.

\section{Results}
In this study, PSO algorithm is used to search the space of 100 × 100 nodes to find the optimal path so the swarm can travel from its initial position to the target point. The investigation of the PSO parameters effect on algorithm performance is shown by the simulation result in Table 1.

\begin{table}[H]
	\centering
	\caption{PSO Parameter and Algorithm Performance}
	\begin{tabular}{@{}cccccc@{}}
		\toprule
		{Initial Position}   &{N Particles} &{$w$} &{$c_1$} &{$c_2$} &{Steps/Iterations} \\ \midrule
		\multirow{2}{*}{(-70, 80)}  & 25                   & 0.90       & 0.60        & 0.65        & 80-120         \\ \cmidrule(l){2-6} 
		& 50                   & 0.80       & 0.65        & 0.90        & 48-58          \\ \midrule
		\multirow{2}{*}{(-80, -10)} & 25                   & 0.60       & 0.90        & 1.45        & 84-133         \\ \cmidrule(l){2-6} 
		& 50                   & 0.45       & 1.20        & 1.70        & 40-56          \\ \bottomrule
	\end{tabular}
\end{table}

The simulation shows that, the larger the population of particles, the fewer steps needed to reach convergence. Even so, the number of particles should be adjusted to the complexity of the obstacles. If the number of particles is too small, the swarm will have difficulty finding a solution due to minimal "communication" between particles during exploration. If the number of particles is too large, the time required to converge to the target will be longer due to too much random motion, but with the optimal number of particles and parameters, the algorithm can work efficiently.

The combination of the weight constant ($w$), particle constant ($c_1$) and global constant ($c_2$) is very important so that the search process becomes more optimal. If these constants are enlarged, the particles will spread easily so that the search process is faster but takes a long time to converge or to "summon" other particles to gather at the target. However, if the constant is too small, the particles will move closely so it will be difficult to find an alternative path if everything is stuck in the local minima. Examples of solution paths obtained from several simulation cases are shown in Figure 3.

Apart from the algorithm, simulation environment needs to be enhance because in some cases particles can penetrating the obstacles due to the velocity being too large. The algorithm above only refuses the particle position update if the particle is not in the free-moving area or is inside of the obstacles coordinates, but if the particle velocity is very fast, the particle displacement becomes larger and the program does not detect it, so particles possible to penetrates the obstacles wall. The solution to this problem is to increase the size of the obstacles wall or limit the particle velocity by setting the optimal constants.

\begin{figure}[H]
	\centering
	{\includegraphics[width=6.5cm]{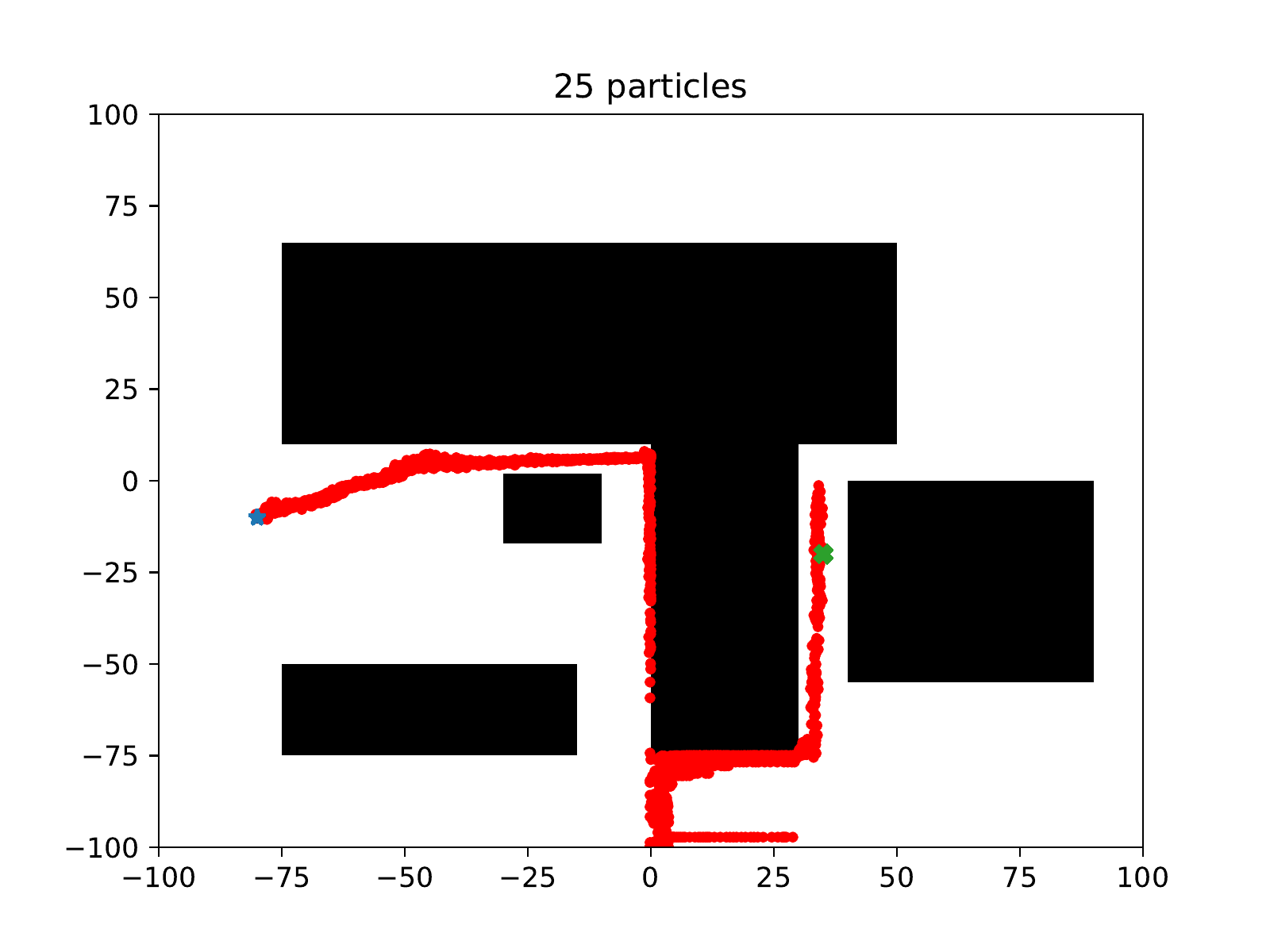}}\hspace{-1.5em}
	{\includegraphics[width=6.5cm]{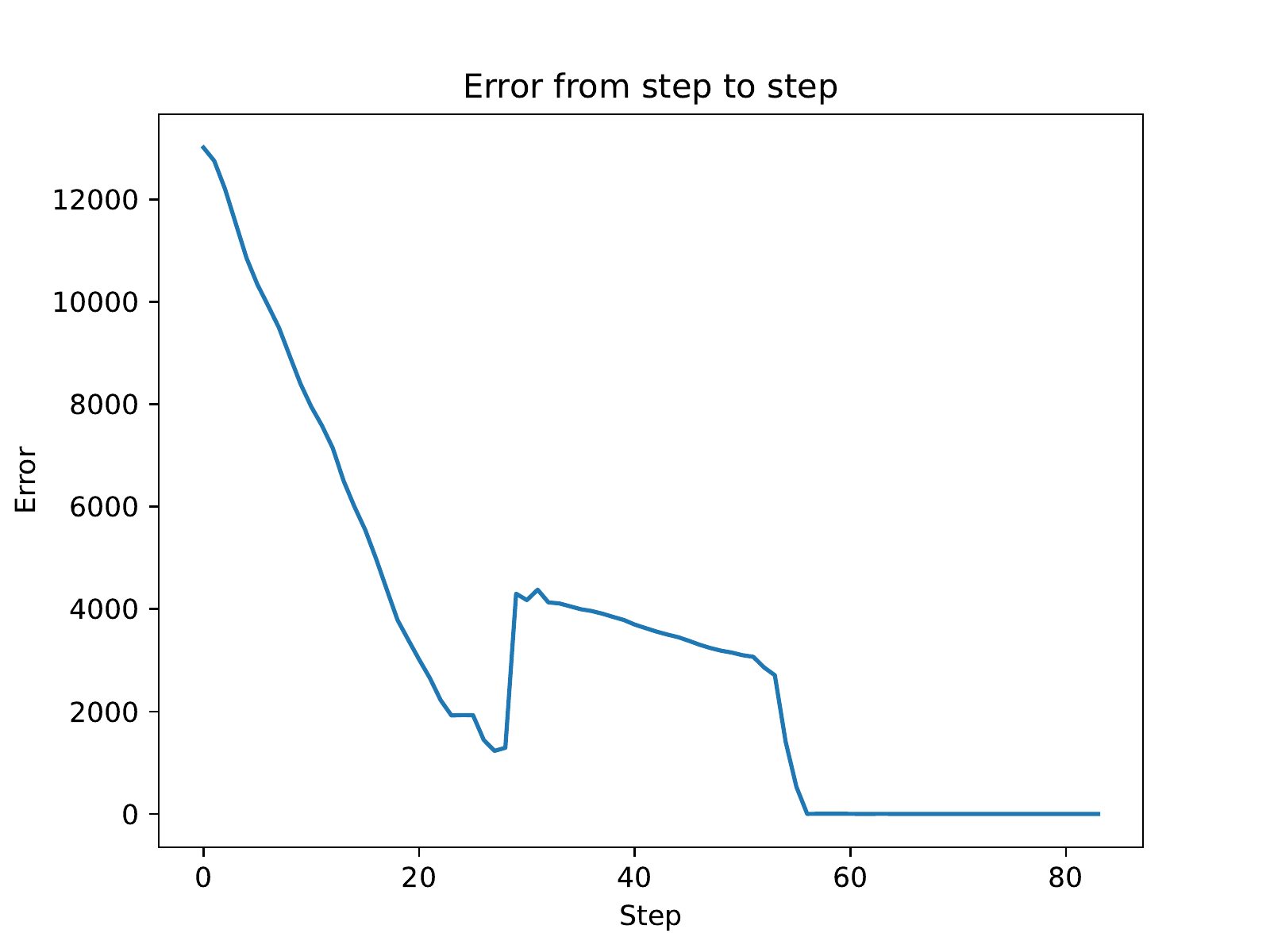}}\vspace{-0.5em}
\end{figure}

\begin{figure}[H]
	\vspace{-2.5em}
	\centering
	{\includegraphics[width=6.5cm]{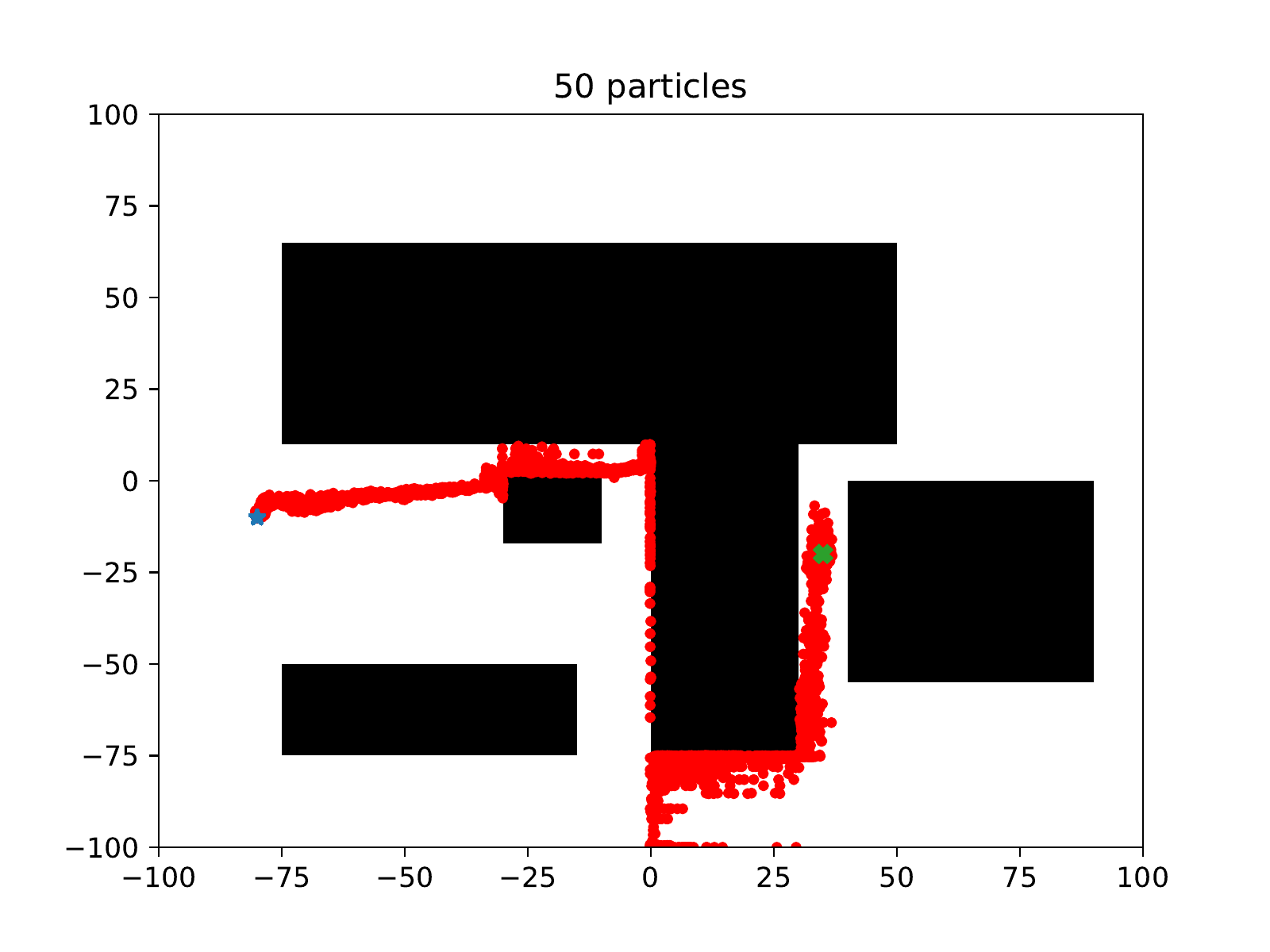}}\hspace{-1.5em}
	{\includegraphics[width=6.5cm]{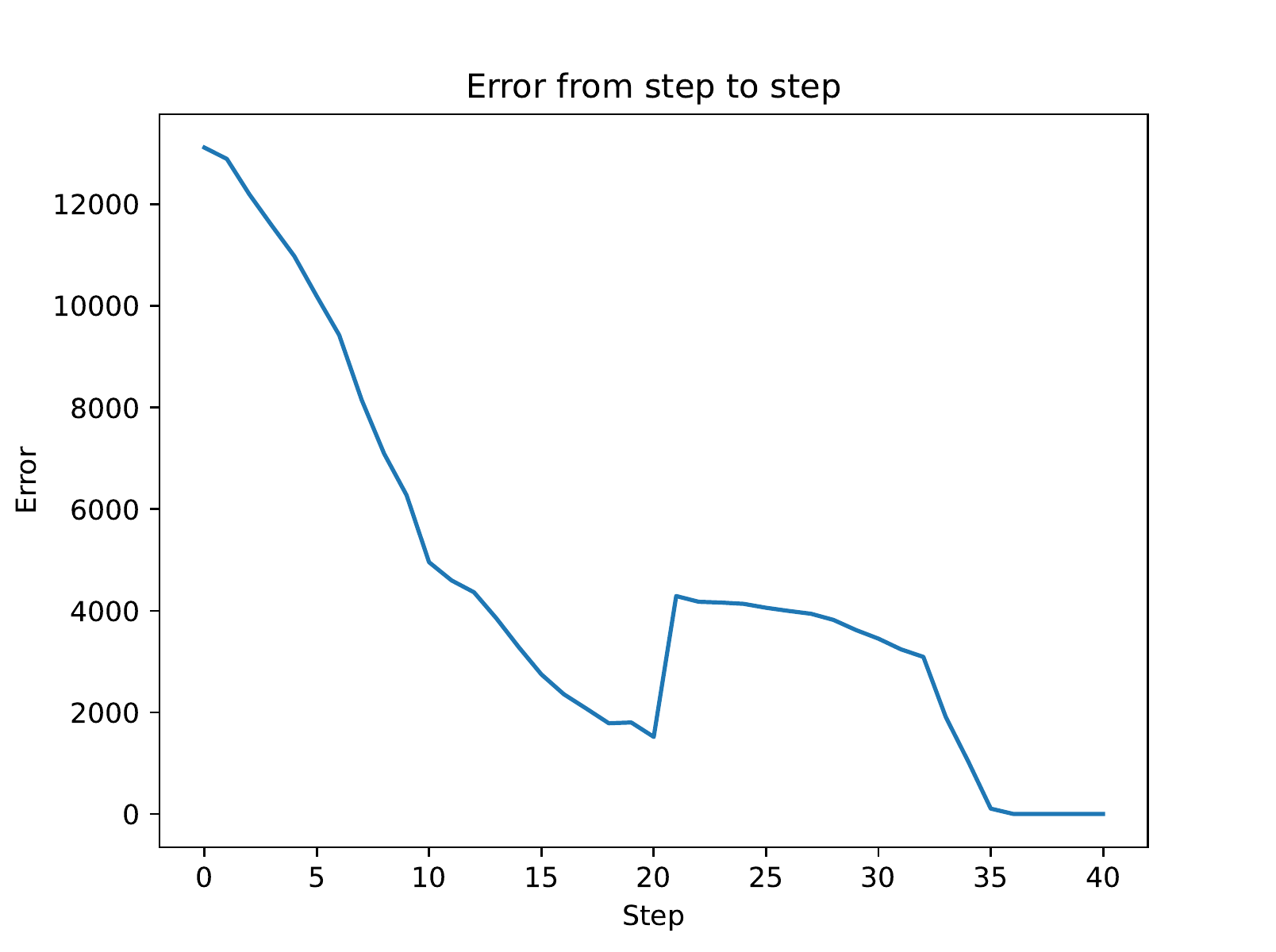}}\vspace{-0.5em}
\end{figure}

\begin{figure}[H]
	\vspace{-2.5em}
	\centering
	{\includegraphics[width=6.5cm]{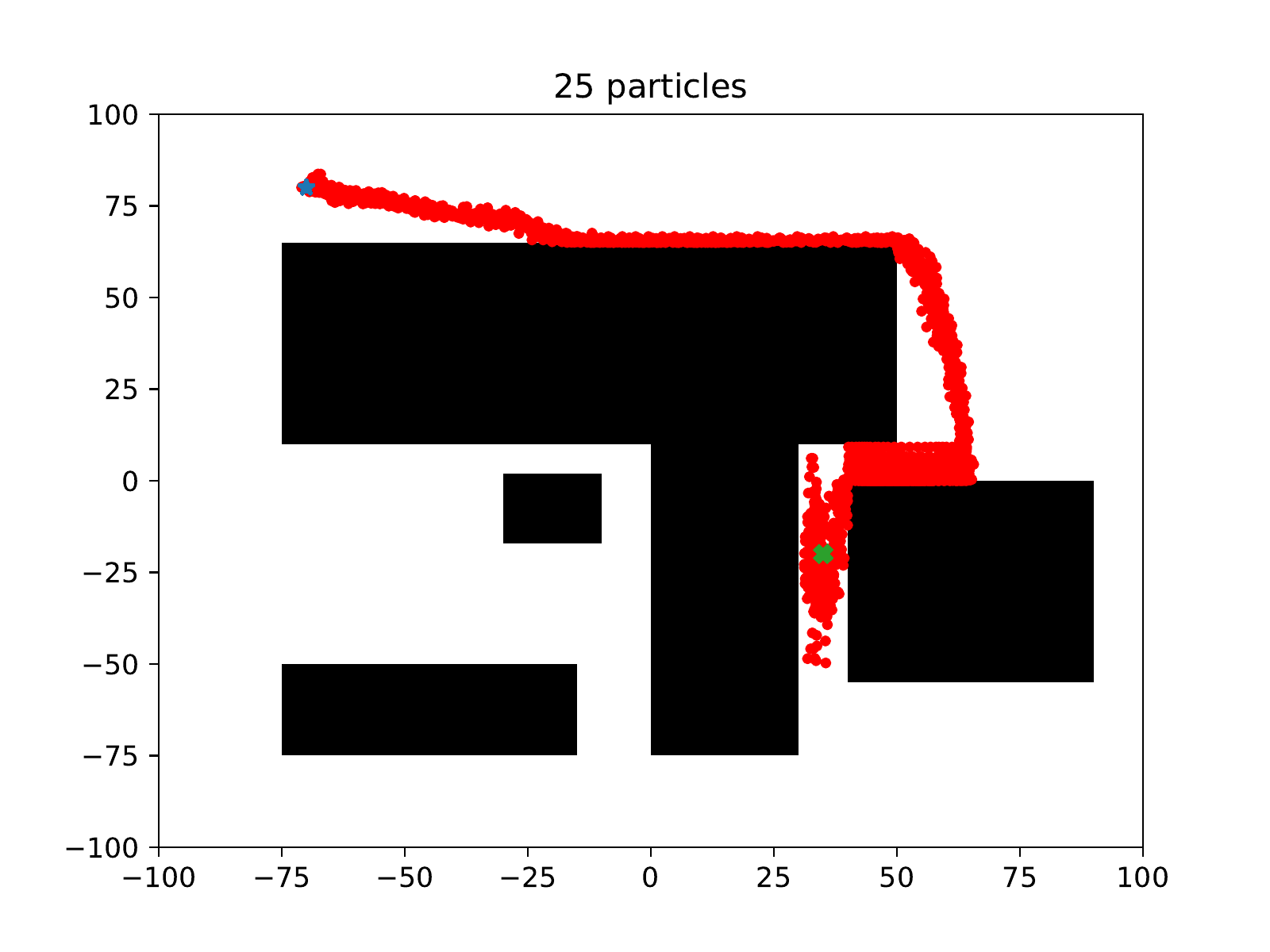}}\hspace{-1.5em}
	{\includegraphics[width=6.5cm]{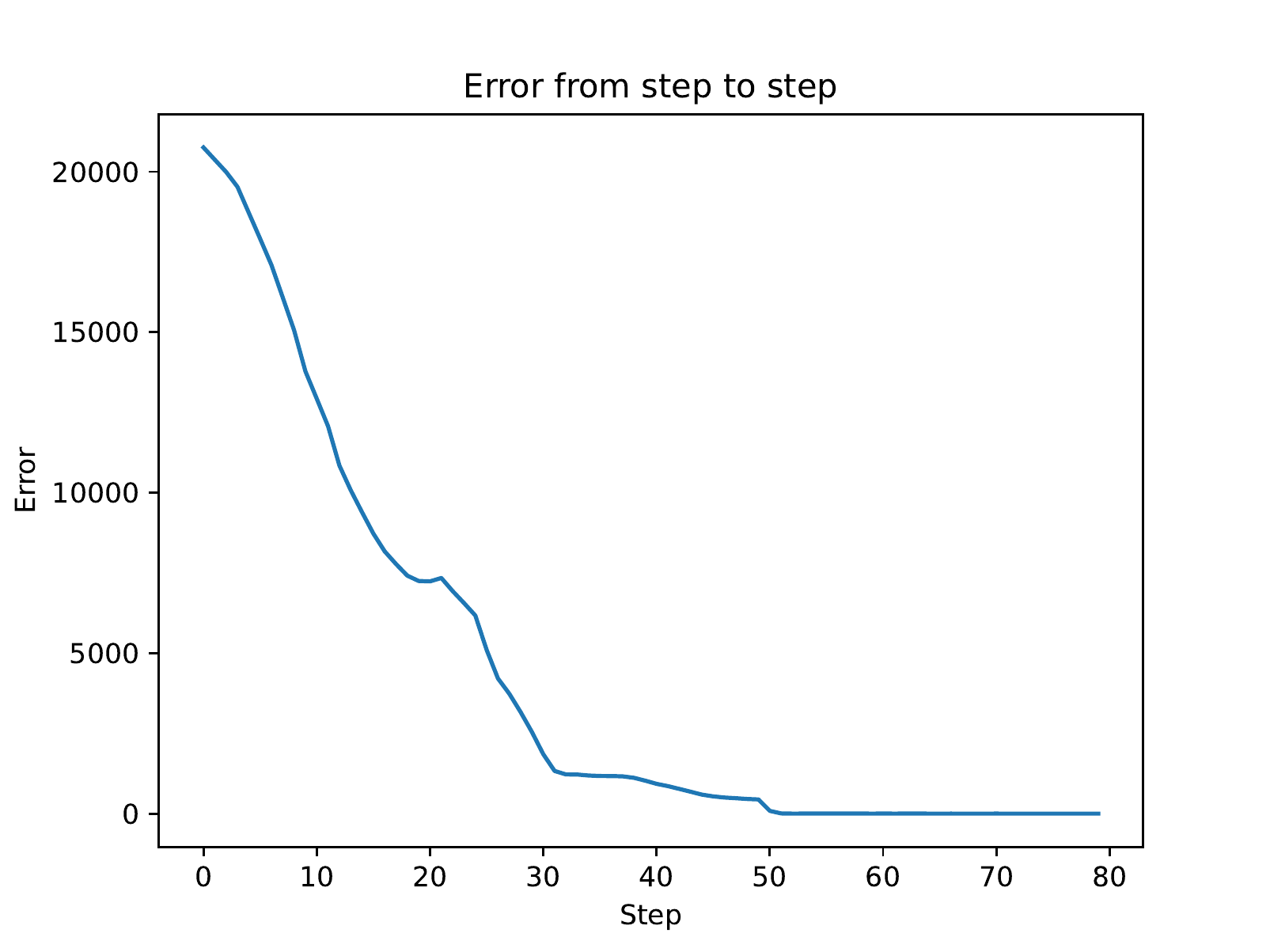}}\vspace{-0.5em}
\end{figure}

\begin{figure}[H]
	\vspace{-2.5em}
	\centering
	{\includegraphics[width=6.5cm]{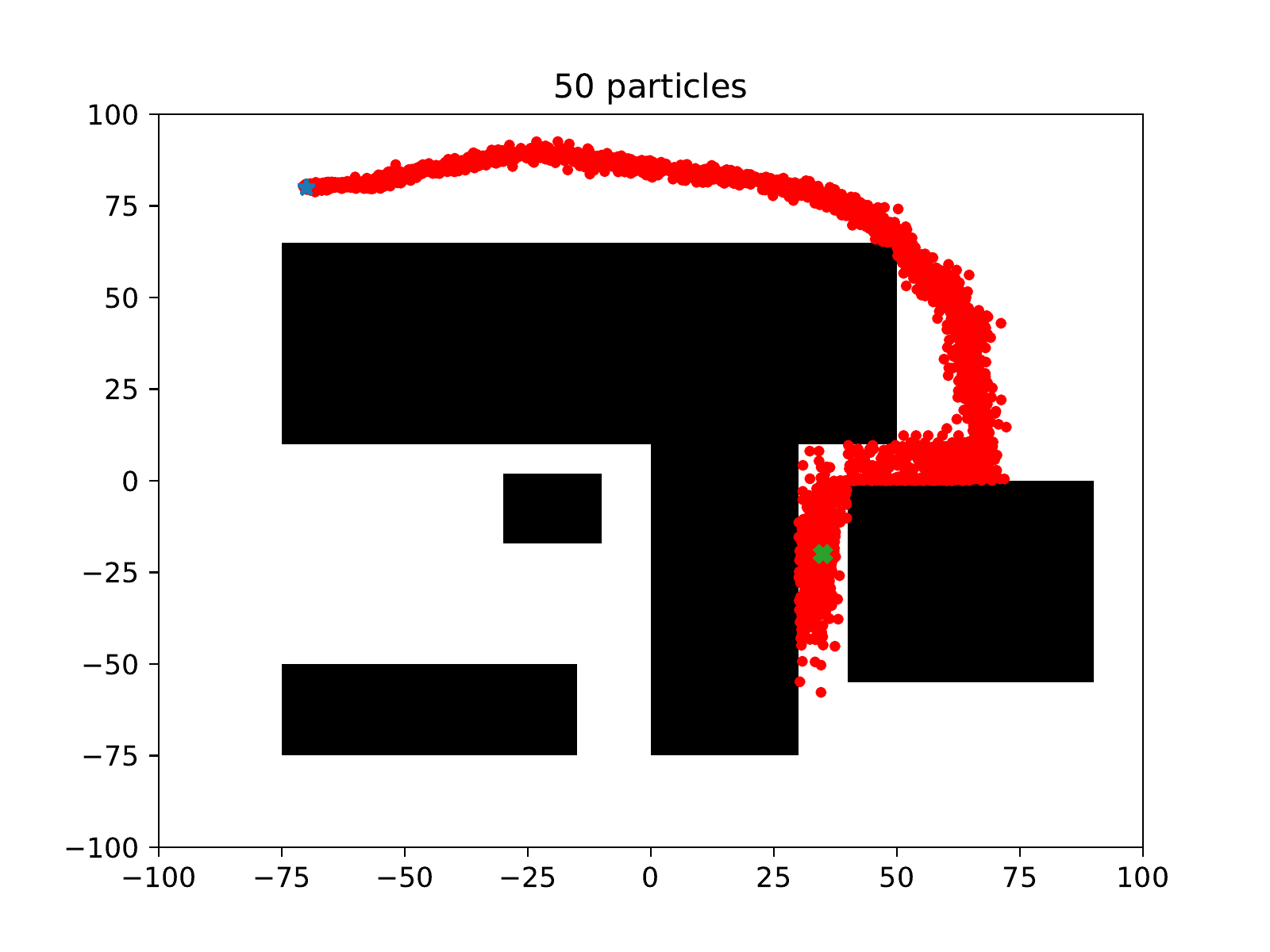}}\hspace{-1.5em}
	{\includegraphics[width=6.5cm]{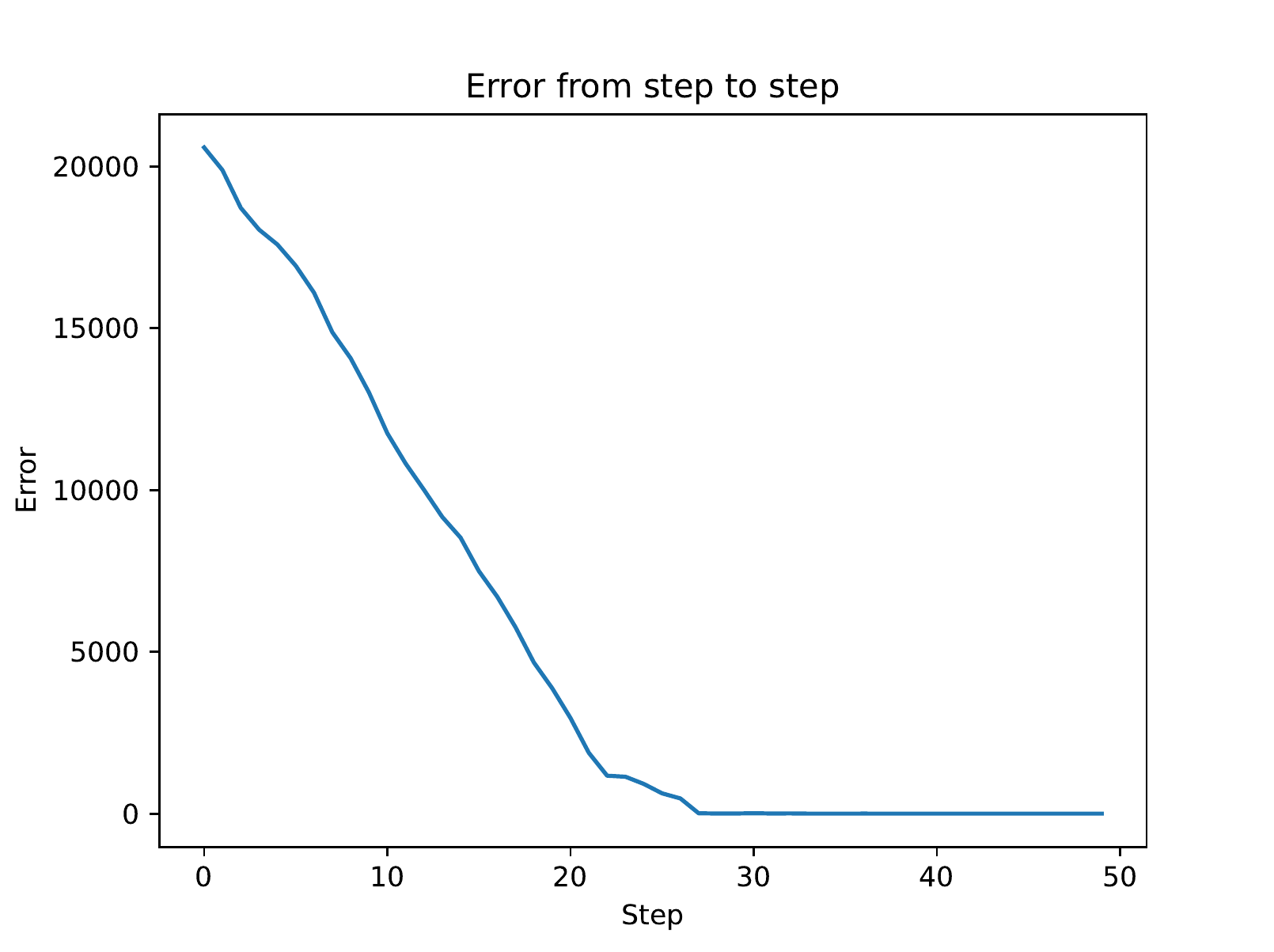}}\vspace{-1em}
	\caption{The particle trace and the error for each step for \textbf{(a,b)} $n$ = 25, ($x_0,y_0$) = (-80,-10); \\ \textbf{(c,d)} $n$ = 50, ($x_0,y_0$) = (-80,-10); \textbf{(e,f)}$n$ = 25, ($x_0,y_0$) = (-70,80); \textbf{(g,h)} $n$ = 50, ($x_0,y_0$) = (-70,80)}
\end{figure}

\section{Conclusion}
This paper presented the algorithm of using the PSO approach to solve the path planning problem. This study also investigated the performance of the evolutionary process with the various parameter and number of particles. Based on the simulations, the PSO algorithm can provide a solution path to different types of tasks in a space with obstacles.

\section*{Code availability}
The implementation code of this research is available for download at \href{https://github.com/dxid-dev/pathfinder}{github.com/dxid-dev/pathfinder}. This research has used \href{https://www.python.org/}{python} and matplotlib \citep{Hunter:2007}.

\section*{Disclosure statement}
No potential conflict of interest was reported by the author.


\begin{thebibliography}{}
\bibitem[Bansal(2019)]{bansal2019particle}Bansal, J. C. (2019). Particle swarm optimization. In \textit{Evolutionary and swarm intelligence algorithms} (bll 11–23). Springer.

\bibitem[Chen et~al.(2020)]{chen2020optimization}Chen, Z., Li, X., Zhu, Z., Zhao, Z., Wang, L., Jiang, S., \& Rong, Y. (2020). The optimization of accuracy and efficiency for multistage precision grinding process with an improved particle swarm optimization algorithm. \textit{International Journal of Advanced Robotic Systems, 17}(1), 1729881419893508.

\bibitem[Falcó et~al.(2020)]{falco2020path}Falcó, A., Hilario, L., Montés, N., Mora, M. C., \& Nadal, E. (2020). A Path Planning Algorithm for a Dynamic Environment Based on Proper Generalized Decomposition. \textit{Mathematics, 8}(12), 2245.

\bibitem[Hunter(2007)]{Hunter:2007}Hunter, J. D. (2007). Matplotlib: A 2D graphics environment. \textit{Computing in Science \& Engineering, 9}(3), 90–95. doi:10.1109/MCSE.2007.55

\bibitem[Karkalos et~al.(2019)]{karkalos2019swarm}Karkalos, N. E., Markopoulos, A. P., \& Davim, J. P. (2019). Swarm intelligence-based methods. In \textit{Computational Methods for Application in Industry 4.0} (bll 33–55). Springer.

\bibitem[Kennedy et~al.(1995)]{kennedy1995particle} Kennedy, J., \& Eberhart, R. (1995). Particle swarm optimization. \textit{Proceedings of ICNN’95-international conference on neural networks, 4}, 1942–1948. IEEE.

\bibitem[Waluyo et~al.(2010)]{psosafril}Waluyo, S. R., Hariadi, M., \& Purnama, I. K. E. (2010). Pencarian Jalur Terbaik Menggunakan Particle Swarm Optimization untuk Mengoptimasi Lalu Lintas Kendaraan.

\bibitem[Wang et~al.(2020)]{wang2020path}Wang, X., Shi, H., \& Zhang, C. (2020). Path planning for intelligent parking system based on improved ant colony optimization. \textit{IEEE Access, 8}, 65267–65273.

\bibitem[Zhu et~al.(2021)]{zhu2021new}Zhu, D.-D., \& Sun, J.-Q. (2021). A new algorithm based on Dijkstra for vehicle path planning considering intersection attribute. \textit{IEEE Access, 9}, 19761–19775.

\end{thebibliography}
\end{document}